\documentclass[11pt,a4paper]{article}
\usepackage[nohyperref]{aacl-ijcnlp2020}
\usepackage{graphicx}
\usepackage{times}
\usepackage{latexsym}
\usepackage{amssymb}
\usepackage{booktabs}
\usepackage{fancyvrb}
\usepackage{multirow}

\usepackage{microtype}

\aclfinalcopy 
\def\aclpaperid{37} 

\title{Stronger Baselines for Grammatical Error Correction \\ Using a Pretrained Encoder--Decoder Model}

\author{Satoru Katsumata\thanks{\ \ Currently working at Retrieva, Inc.} \and Mamoru Komachi\\
  Tokyo Metropolitan University \\
  \texttt{satoru.katsumata@retrieva.jp, komachi@tmu.ac.jp}
  }

\date{}

\begin{document}
\maketitle
\begin{abstract}
    Studies on grammatical error correction (GEC) have reported the effectiveness of pretraining a Seq2Seq model with a large amount of pseudodata.
    However, this approach requires time-consuming pretraining for GEC because of the size of the pseudodata.
    In this study, we explore the utility of bidirectional and auto-regressive transformers (BART) as a generic pretrained encoder--decoder model for GEC.
    With the use of this generic pretrained model for GEC, the time-consuming pretraining can be eliminated.
    We find that monolingual and multilingual BART models achieve high performance in GEC, with one of the results being comparable to the current strong results in English GEC.
    Our implementations are publicly available at GitHub\footnote{https://github.com/Katsumata420/generic-pretrained-GEC}.
\end{abstract}

\section{Introduction}
Grammatical error correction (GEC) is the automatic correction of grammatical and other language-related errors in text.
Most works regard this task as a translation task and use encoder--decoder (Enc--Dec) architectures to convert ungrammatical sentences to grammatical ones.
This Enc--Dec approach often does not require linguistic knowledge of the target language.
Strong Enc--Dec models for GEC are pretrained with a large amount of artificially generated data, commonly referred to as `pseudodata', that is created by introducing artificial error to a monolingual corpus.
Hereafter, pretraining using pseudodata aimed at the GEC task is referred to as \textbf{task-oriented} pretraining \cite{kiyono2019,beasota,low_resource_gec,kaneko_bert}.
For example, \citet{kiyono2019} generated a pseudo corpus using back-translation and achieved strong results for English GEC.
\citet{low_resource_gec} generated a pseudo corpus by introducing artificial errors into monolingual corpora and achieved the best scores for GEC in several languages by adopting the methods proposed by \citet{beasota}.

These task-oriented pretraining approaches require extensive use of a pseudo-parallel corpus.
Specifically, \citet{beasota} used 100M ungrammatical and grammatical sentence pairs, while \citet{kiyono2019} and \citet{kaneko_bert} used 70M sentence pairs,
 which required time-consuming pretraining of GEC models using the pseudo corpus.

In this study, we determined the effectiveness of publicly available pretrained Enc--Dec models for GEC.
Specifically, we investigated pretrained models without the need for pseudodata.
We explored a pretrained model proposed by \citet{bart} called bidirectional and auto-regressive transformers (BART).
\citet{m-bart} also proposed multilingual BART.
These models were pretrained by predicting the original sequence, given a masked and shuffled sentence.
The motivation for using these models for GEC was that it achieved strong results for several text generation tasks, such as summarization;
 we refer to it as a \textbf{generic} pretrained model.

We used generic pretrained BART models to compare with GEC models using a pseudo-corpus approach \cite{kiyono2019,kaneko_bert,low_resource_gec}.
We conducted GEC experiments for four languages: English, German, Czech, and Russian.
The Enc--Dec model based on BART achieved results comparable with those of current strong Enc--Dec models for English GEC.
The multilingual model also showed high performance in other languages, despite only requiring fine-tuning.
These results suggest that BART can be used as a simple baseline for GEC.

\section{Previous Work}
The Enc--Dec approach for GEC often uses the task-oriented pretraining strategy.
For example, \citet{copy_gec} and \citet{beasota} reported that pretraining of the Enc--Dec model using a pseudo corpus is effective for the GEC task.
In particular, they introduced word- and character-level errors into a sentence in monolingual corpora.
They developed a confusion set derived from a spellchecker and randomly replaced a word in a sentence.
They also randomly deleted a word, inserted a random word, and swapped a word with an adjacent word.
They performed these same operations, i.e., replacing, deleting, inserting, and swapping, for characters.
The pseudo corpus made by the above methods consisted of 100M training samples.
Our study aims to investigate whether the generic pretrained models are effective for GEC,
 because pretraining with such a large corpus is time-consuming.

\citet{low_resource_gec} adopted \citet{beasota}'s method for several languages, including German, Czech, and Russian.
They trained a Transformer \cite{attention} with pseudo corpora (10M sentence pairs), and
 achieved current state-of-the-art (SOTA) results for German, Czech, and Russian GEC.
We compared their results with those of the generic pretrained model
to confirm whether the model was effective for GEC in several languages.

\citet{kiyono2019} explored the generation of a pseudo corpus by introducing random errors or using back-translation.
They reported that a task-oriented pretraining with back-translation data and character errors is better than that with pseudodata based on random errors.
\citet{kaneko_bert} combined \citet{kiyono2019}'s pretraining approach with BERT \cite{bert2019} and improved \citet{kiyono2019}'s results.
Specifically, \citet{kaneko_bert} fine-tuned BERT with a grammatical error detection task.
The fine-tuned BERT outputs for each token were combined with the original tokens as a GEC input.
Their study is similar to our research
 in that both studies use publicly available generic pretrained models to perform GEC.
The difference between these studies is that \citet{kaneko_bert} used the architecture of the pretrained model as an encoder.
Therefore, their method still requires pretraining with a large amount of pseudodata.

The current SOTA approach for English GEC uses the sequence tagging model proposed by \citet{gector}.
They designed token-level transformations to map input tokens to target corrections to produce training data.
The sequence tagging model then predicts the transformation corresponding to the input token.
We do not attempt to make a comparison with this approach,
 as the purpose of our study is to create a strong GEC model without using pseudodata or linguistic knowledge.

\section{Generic Pretrained Model}
BART \cite{bart} is pretrained by predicting an original sequence, given a masked and shuffled sequence using a Transformer.
They introduced masked tokens with various lengths based on the Poisson distribution, inspired by SpanBERT \cite{spanbert}, at multiple positions.
BART is pretrained with large monolingual corpora (160 GB), including news, books, stories, and web-text domains.
This model achieved strong results in several generation tasks; thus, it is regarded as a generic model.

They released pretrained models using English monolingual corpora for several tasks, including summarization, which we used for English GEC.
\citet{m-bart} proposed multilingual BART (mBART) for a machine translation task, which we used for GEC of several languages.
The latter model was trained using monolingual corpora for 25 languages simultaneously.
They used a special token for representing the language of a sentence.
For example, they added \verb|<de_DE>| and \verb|<ru_RU>| into the initial token of the encoder and decoder for De--Ru translation.
To fine-tune mBART for German, Czech, and Russian GEC, we set the target language for the special token referring to that language.

\section{Experiment}
\subsection{Settings}

\begin{table}[t]
    \centering
    \small
    \begin{tabular}{llrrr} \toprule
        lang & Corpus & Train & Dev & Test \\ \midrule \midrule
                 & BEA & 1,157,370 & 4,384 & 4,477 \\
        En       & JFLEG & - & - & 747 \\
           & CoNLL-2014 & - & - & 1,312 \\ \midrule
        De & Falko+MERLIN & 19,237 & 2,503 & 2,337 \\ \midrule
        Cz & AKCES-GEC & 42,210 & 2,485 & 2,676 \\ \midrule
        Ru & RULEC-GEC & 4,980 & 2,500 & 5,000 \\ \bottomrule
    \end{tabular}
    \caption{Data statistics.}
    \label{gec-data}
\end{table}

\begin{table*}[t]
  \centering
  \small
  \begin{tabular}{lrrrrrrr} \toprule
      &  \multicolumn{3}{c}{CoNLL-14 ($\mathrm{M^2}$)} & JFLEG & \multicolumn{3}{c}{BEA-test} \\ \cmidrule(lr){2-4} \cmidrule(lr){5-5} \cmidrule(lr){6-8}
      & P & R & $\mathrm{F_{0.5}}$ & GLEU &  P & R & $\mathrm{F_{0.5}}$   \\ \midrule
      \citet{kiyono2019} & 67.9/\underline{73.3} & 44.1/44.2 & 61.3/64.7 & 59.7/61.2 & 65.5/\underline{74.7} & 59.4/56.7 & 64.2/\underline{70.2} \\
      \citet{kaneko_bert} & 69.2/72.6 & \textbf{45.6}/\underline{46.4} & \textbf{62.6}/\underline{65.2} & \textbf{61.3}/\underline{62.0} & 67.1/72.3 & \textbf{60.1}/\underline{61.4} & \textbf{65.6}/69.8 \\
      BART-based & \textbf{69.3}/69.9 & 45.0/45.1 & \textbf{62.6}/63.0 & 57.3/57.2 & \textbf{68.3}/68.8 & 57.1/57.1 & \textbf{65.6}/66.1 \\
      \bottomrule
  \end{tabular}
  \caption{English GEC results. Left and right scores represent single and ensemble model results, respectively. Bold scores represent the best score in the single models, and underlined scores represent the best overall score.}
  \label{result_score_english}
\end{table*}

\paragraph{Common Settings.}
As presented in Table \ref{gec-data}, we used learner corpora, including
BEA\footnote{BEA corpus is made of several corpora. Details can be found in \citet{bea2019}.} \cite{bea2019,locness,lang8-1,lang8-2,fce,nucle}, JFLEG \cite{jfleg}, and CoNLL-14 \cite{ng2014} data for English; Falko+MERLIN data \cite{merlin} for German; AKCES-GEC \cite{low_resource_gec} for Czech; and RULEC-GEC \cite{rulec} for Russian.

Our models were fine-tuned using a single GPU (NVIDIA TITAN RTX), and
 our implementations were based on publicly available code\footnote{BART, mBART: https://github.com/pytorch/fairseq}.
 We used the hyperparameters provided in some previous works \cite{bart,m-bart}, unless otherwise noted.

The scores excluding the ensemble method were averaged in five fine-tuned experiments with random seeds.

\paragraph{English.}
Our setting for the English datasets was almost the same as that of \citet{kiyono2019}.
We extracted the training data from BEA-train for English GEC.
Similar to \citet{kiyono2019}, we did not use the unchanged sentences in the source and target sides; thus, the training data consisted of 561,525 sentences.
We used BEA-dev to determine the best model.

We trained the BART-based models by using \verb|bart.large|.
This model was proposed for the summarization task, which required some constraints in inference to ensure appropriate outputs;
 however, we did not impose any constraints because our task was different.
We applied byte pair encoding (BPE) \cite{bpe} to the training data for the BART-based model by using the BPE model of \citet{bart}.

We used the $\mathrm{M^2}$ scorer \cite{m2score} and GLEU \cite{gleu} for CoNLL-14 and JFLEG, respectively, and
used the ERRANT scorer \cite{errant} for BEA-test.
We compared these scores with strong results \cite{kiyono2019,kaneko_bert}.

\begin{table}[t]
    \centering
    \scalebox{0.8}{
    \begin{tabular}{clrrr} \toprule
        & & P & R & $\mathrm{F_{0.5}}$ \\ \midrule
        \multirow{2}{*}{De} & \citet{low_resource_gec} & 78.21 & 59.94 & 73.31 \\
           & mBART-based & 73.97 & 53.98 & 68.86 \\ \midrule
        \multirow{2}{*}{Cz} & \citet{low_resource_gec} & 83.75 & 68.48 & 80.17 \\
         & mBART-based & 78.48 & 58.70 & 73.52 \\ \midrule
           & \citet{low_resource_gec} & 63.26 & 27.50 & 50.20 \\
        Ru & mBART-based & 32.13 & 4.99 & 15.38 \\
           & \ \ with pseudo corpus & 53.50 & 26.35 & 44.36 \\ \bottomrule
    \end{tabular}
}
    \caption{German, Czech, and Russian GEC results. These models are not an ensemble of multiple models.}
    \label{result_score_others}
\end{table}

\paragraph{German, Czech, and Russian.}
The dataset settings in this study were almost the same as those used by \citet{low_resource_gec} for each language.
We used official training data and decided the best model by using the development data.

In addition, we trained the mBART-based models for German, Czech, and Russian GEC.
We used \verb|mbart.cc25| for the mBART-based models.
For the mBART-based model, we followed \citet{m-bart};
 we detokenized\footnote{We used detokenizer.perl in the Moses script \cite{moses}.} the GEC training data for the mBART-based model and applied SentencePiece \cite{spm} with the SentencePiece model shared by \citet{m-bart}.
Using this preprocessing, the input sentence may not represent grammatical information, compared with the sentence tokenized using a morphological analysis tool and subword tokenizer.
However, what preprocessing is appropriate for GEC is beyond this paper's scope and will be treated as future work.
For evaluation, we tokenized the outputs after recovering the subwords.
Then, we used a spaCy-based\footnote{https://spacy.io} tokenizer for German\footnote{We used the built-in de model.} and Russian\footnote{https://github.com/aatimofeev/spacy\_russian\_tokenizer}, and the MorphoDiTa tokenizer\footnote {https://github.com/ufal/morphodita} for Czech.

Moreover, the $\mathrm{M^2}$ scorer was used for each language.
We compared these scores with the current SOTA results \cite{low_resource_gec}.

\subsection{Results}
\paragraph{English.}
Table \ref{result_score_english} presents the results of the English GEC task.
When using a single model, the BART-based model is better than the model proposed by \citet{kiyono2019}, and the results are comparable to those reported by \citet{kaneko_bert} in terms of CoNLL-14 and BEA-test.
\citet{kiyono2019} and \citet{kaneko_bert} incorporated several techniques to improve the accuracy of GEC.
To compare these models, we experimented with an ensemble of five models.
Our ensemble model was slightly better than our single model, but worse than the ensemble models by \citet{kiyono2019} and \citet{kaneko_bert}.
The BART-based model along with the ensemble model achieved results comparable to current strong results despite only requiring fine-tuning of the BART model.
We believe that the reason for the ineffectiveness of the ensemble method is that the five models are not significantly different
 as the initial weights are the same as those of the BART model,
 and seeds only affect minor changes, such as training data order, and so on.

\paragraph{German, Czech, and Russian.}
Table \ref{result_score_others} presents the results for German, Czech, and Russian GEC.

In the German GEC task, the mBART-based model achieves 4.45 $\mathrm{F_{0.5}}$ points lower than the model by \citet{low_resource_gec}.
This may be because \citet{low_resource_gec} pretrains the GEC model with only the target language,
 whereas mBART is pretrained with 25 languages, resulting in
 the information of other languages being included as noise.

In the Czech GEC task, the mBART-based model achieves 6.65 $\mathrm{F_{0.5}}$ points lower than the model by \citet{low_resource_gec}.
Similar to the case of the German GEC results, we suppose that mBART includes noisy information.

Considering Russian GEC, the mBART-based model shows much lower scores than \citet{low_resource_gec}'s model.
This may be because the training data for
 Russian GEC are scarce compared to those of German or Czech.
To investigate the effect of corpus size, we additionally trained the mBART model with a 10M pseudo corpus, using the method proposed by \citet{beasota}, and fine-tuned it with the learner corpus to compensate for the low-resource scenario.
The results presented in Table \ref{result_score_others} support our hypothesis.

\begin{table}[t]
    \centering
    \small
    \begin{tabular}{lrrrrrr} \toprule
          & \multicolumn{3}{c}{\citet{kaneko_bert}} &
          \multicolumn{3}{c}{BART-based} \\ \cmidrule(lr){2-4} \cmidrule(lr){5-7}
        Error Type   & P & R & $\mathrm{F_{0.5}}$ & P & R & $\mathrm{F_{0.5}}$ \\ \midrule
        PUNCT & 74.1 & 52.7 & 68.5 & 79.2 & 59.0 & \textbf{74.1} \\
        DET & 73.7 & 72.9 & 73.5 & 76.3 & 71.1 & \textbf{75.2} \\
        PREP & 73.4 & 69.1 & \textbf{72.5} & 71.2 & 64.8 & 69.9 \\
        ORTH & 86.9 & 62.9 & \textbf{80.8} & 84.2 & 52.9 & 75.3 \\
        SPELL & 83.1 & 79.5 & \textbf{82.3} & 84.7 & 55.2 & 76.5 \\ \bottomrule
    \end{tabular}
    \caption{BEA-test scores for the top five error types, except for OTHER. \citet{kaneko_bert} and BART-based are ensemble models. Bold scores represent the best score for each error type.}
    \label{analysis_bart}
\end{table}

\section{Discussion}
\paragraph{BART as a simple baseline model.}
According to the German and Czech GEC results, the mBART-based model, in which we only fine-tuned the pretrained mBART model, achieves comparable scores with SOTA models.
In other words, mBART-based models are considered to show sufficiently high performance for several languages without using a pseudo corpus.
These results indicate that the mBART-based model can be used as a simple GEC baseline for several languages.

\paragraph{Performance comparison for each error type.}
We compare the BART-based model with \citet{kaneko_bert}'s model for common error types using a generic pretrained model.
Table \ref{analysis_bart} presents the results for the top five error types in BEA-test.
According to these results, BART-based is superior to \citet{kaneko_bert} in PUNCT and DET errors;
 in particular, PUNCT is 5.6 $\mathrm{F_{0.5}}$ points better.
BART is pretrained to correct the shuffled and masked sequence,
 so that this model learns to place punctuation adequately.
In contrast, \citet{kaneko_bert} uses an encoder
 that is not pretrained with correcting shuffled sequences.

Conversely, \citet{kaneko_bert} report better results for other errors, except for DET.
Regarding ORTH and SPELL, their model is more than 5 $\mathrm{F_{0.5}}$ points better than the BART-based one.
It is difficult for the BART-based model to correct these errors because BART uses shuffled and masked sequences as noise in pretraining; not using character-level errors.
\citet{kaneko_bert} introduce character errors into a pseudo corpus as task-oriented Enc--Dec pretraining;
 this is the reason why the BART-based model is inferior to \citet{kaneko_bert} in these errors.

\section{Conclusion}
We introduced a generic pretrained Enc--Dec model, BART, for GEC.
The experimental results indicated that BART better initialized the Enc--Dec model parameters.
The fine-tuned BART achieved remarkable results, which were comparable to the current strong results in English GEC.
Indeed, the monolingual BART seems to be more effective for GEC than the model with a multilingual setting.
However, although it is not as good as SOTA, fine-tuned mBART exhibited high performance in other languages.
This implies that
 BART is a simple baseline model for pretraining GEC methods because it only requires fine-tuning as training.

\section*{Acknowledgements}
We thank the anonymous reviewers for their insightful comments.
This work has been partly supported by the programs of the
Grant-in-Aid for Scientific Research from the Japan Society for the
Promotion of Science (JSPS KAKENHI) Grant Numbers 19K12099 and
19KK0286.

\bibliography{ref}
\bibliographystyle{acl_natbib}

\end{document}